%% file: IEEEtran.tex
\documentclass[conference]{IEEEtran}
\IEEEoverridecommandlockouts
\usepackage{cite}
\usepackage{amsmath,amssymb,amsfonts}
\usepackage{algorithmic}
\usepackage{graphicx}
\usepackage{textcomp}
\usepackage{xcolor}
\usepackage{multirow} 
\usepackage{microtype}
\usepackage{subcaption}
\usepackage{tabularx}
\usepackage{lipsum}
\usepackage{latexsym}
\usepackage{times}

\def\BibTeX{{\rm B\kern-.05em{\sc i\kern-.025em b}\kern-.08em
    T\kern-.1667em\lower.7ex\hbox{E}\kern-.125emX}}
\begin{document}

\title{Mix of Experts Language Model for Named Entity Recognition \\
}

\author{
\IEEEauthorblockN{Xinwei Chen}
\IEEEauthorblockA{\textit{Department of Electronical and Computer Engineering} \\
\textit{University of Illinois at Urbana Champaign}\\
Urbana, USA\\
xinweic2@illinois.edu }
\and

\IEEEauthorblockN{Kun Li}
\IEEEauthorblockA{\textit{Department of Computer Science} \\
\textit{University of Illinois at Urbana Champaign}\\
Urbana, USA \\
kunli3@illinois.edu}

\and

\IEEEauthorblockN{Jiangjian Guo}
\IEEEauthorblockA{\textit{Department of Computer Science and Engineering} \\
\textit{University of California San Diego}\\
La Jolla, US \\
j9guo@ucsd.edu }

\and

\IEEEauthorblockN{Tianyou Song}
\IEEEauthorblockA{\textit{Department of Computer Science} \\
\textit{Columbia University}\\
New York, USA \\
tianyou.song@columbia.edu}
}

\maketitle

\input{0_abstract.tex}
\input{1_Introduction.tex}
\input{2_Related_work.tex}
\input{3_Preliminaries}
\input{4_Model}
\input{5_Experiment}
\input{6_Conclusion}

\bibliographystyle{IEEEtran}
\bibliography{custom}
\end{document}

%% file: 0_abstract.tex
\begin{abstract}
    Named Entity Recognition (NER) is an essential steppingstone in the field of natural language processing. Although promising performance has been achieved by various distantly supervised models, we argue that distant supervision inevitably introduces incomplete and noisy annotations, which may mislead the model training process. To address this issue, we propose a robust NER model named BOND-MoE based on Mixture of Experts (MoE). Instead of relying on a single model for NER prediction, multiple models are trained and ensembled under the Expectation-Maximization (EM) framework, so that noisy supervision can be dramatically alleviated. In addition, we introduce a fair assignment module to balance the document-model assignment process. Extensive experiments on real-world datasets show that the proposed method achieves state-of-the-art performance compared with other distantly supervised NER. 
\end{abstract}


%% file: 1_Introduction.tex
\section{Introduction}
Named Entity Recognition (NER) is a sequence tagging NLP task which locates and classifies entities from text into pre-defined categories such as organization, persons, and locations. Supervised learning techniques by training models on large, annotated corpus came to play and achieved state-of-the-art performance on NER tasks~\cite{devlin2018bert, chen2024fewshot}. Some effective supervised algorithms such as  Conditional Random Fields (CRF) \cite{10.5555/645530.655813}, bimodal CNN \cite{li-23-deception-detection}. However, traditional supervised methods require heavy human efforts in annotating token-level labels, which is time-consuming in real-world applications. In addition, this dependency on human annotations could introduce errors and biases in NER results. Insufficient labeled data made supervised methods inappropriate for real-world NER tasks. 
In this paper we propose a novel mixed model which incorporates Bert with Mixture of Expert (MoE) for distantly supervised NER task. We test the effectiveness of our method on 5 real world datasets. Our contributions include: 

\begin{itemize}
\item We firstly incorporated the MoE module into a self-training process BOND framework for distantly supervised NER task.
\item We introduce a fair assignment module for documents assignment process.
\end{itemize}

%% file: 2_Related_work.tex
\section{Related Work}\label{sec:related}

Distant supervision is proposed to tackle lack of labeled data. This approach is to match tokens in the corpus with tags in external knowledge base, such as Wikipedia, and even pre-trained language models (PLM) (e.g., GPT-3~\cite{brown2020language}). However, distantly supervised approaches may suffer from incomplete annotations due to limited coverage of tokens by knowledge base, as well as noisy annotation caused by ambiguous label matching. \cite{zeng23c, zeng2023} propose an generative graph dictionary learning to for classification. \cite{li-24-vqa, yang-24-data-aug} use Generative Adversarial Networks (GNN) with attention mechanism for categorizing problems. Among all the works that attempt to solve these two challenges.

\paragraph{BOND} 
BOND \cite{liang2020bond} leverages BERT in a two-stage training process for distant supervision in NER. In the first stage, it fine-tunes BERT using distantly matched labels, incorporating early stopping to avoid over-fitting. The second stage involves self-learning with soft labels generated by the fine-tuned model. BOND effectively addresses noisy annotations in distant supervision.

\paragraph{MoE}
BOND relys solely on a single PLM may not handle noisy or incomplete annotations effectively. In the MoE, a layer comprises K feed-forward neural networks (experts) and a gating network. These experts are trained independently on distinct document subsets, identified using the expectation-maximization (EM) framework. During inference, their outputs are combined, significantly reducing the impact of noisy annotations. Additionally, the MoE generates interpretable topics, as each expert’s clustered representations reveal hidden attributes. Moreover, by assigning different experts to diverse categories within the same named entity, the MoE demonstrates robustness in handling ambiguity label matching.

\paragraph{BOND-MoE}

To address the issues, we propose BOND-MoE, which incorporates PLM BERT together with the MoE module for distantly supervised NER task. which consists of three stages. In the first stage, documents are assigned to different experts for independent training with the hard-EM algorithm. At the expectation step, each input document is fed into the experts with, based on which the expert with the highest posterior probability on average scores is chosen among all tokens in each input document. At maximization step, we train the selected experts based on the assigned documents to maximize corresponding NER losses. In the extreme case, it is possible that all documents are assigned to the same expert for training while other models have no samples. To avoid such circumstance, we propose a fair assignment model for document-expert assignment in the second stage. In the last stage, we incorporate the MoE module into self-training process like BOND framework for distantly supervised NER task. The student expert is trained on generated pseudo labels, then the teacher expert iteratively updates itself and generates a new set of pseudo labels for the next training step of the student expert. 

%% file: 3_Preliminaries.tex
\section{Preliminaries}

\subsection{Problem Definition}

In this section, we will define the problem of Distantly supervised NER.  We use $Y = \{y_1,\dots,y_m\}$ to represent the set of entity types, $D = \{d_1,\dots,d_N\}$ to represent the set of documents where each document is composed of a series tokens $d_i=[t_{i,1},\dots,t_{i,n_{i}}]$ and $M=\{m_1,\dots,m_K\}$ to represent a set of NER models. 

For supervised NER task, we are given a set of documents with well-annotated entity types denoted as $\mathbf{y}_i=[y_{i,1},\dots,y_{i,n_i}]$ for document $d_i$. The goal of supervised NER task is to train a model with parameters $\theta$ to predict the entity label by minimizing the prediction loss as follows~\cite{liang2020bond}:

\begin{equation}
    \mathop{\arg\min}_\theta \frac{1}{N}\sum_{i=1}^Nl(f(d_i;\theta),\mathbf{y}_i)
\end{equation}

However, as stated in the previous section, supervised NER approaches inevitably suffer from the heavy human annotation efforts and the poor generalization ability to new domains. To address these issues, leveraging distant supervision from existing knowledge base and pretrained language models provide a feasible way. Formally speaking, the distantly supervised NER task can be defined as follows: We have a set of entity classes $Y$, a set of documents $D$, and external knowledge bases or language models to predicted entity types $\hat{\mathbf{y}}_i=[\hat{y}_{i,1},\allowbreak\dots,\allowbreak\hat{y}_{i,n_i}]$ for tokens in each document $d_i$.

We also introduce the concept of MoE to alleviate noises in the distant labels. Formally speaking, the MoE task can be defined as follows: We have a set of inputs $X=\{\mathbf{x}_1,\dots,\mathbf{x}_N\}$ and corresponding labels $Y = \{\mathbf{y}_1,\dots,\mathbf{y}_N\}$, to get a set of experts $f_i:\mathbf{x}\to\mathbf{y}$ with parameters $\Theta = \{\theta_1,\dots,\theta_M\}$ such that:
\begin{equation}
    \begin{aligned}
    \max_{\Theta}p(Y|X,\Theta)&=\sum_{i=1}^Mp(Y,i|X,\theta_i)\\
        &=\sum_{i=1}^Mp(\theta_i|X)p(Y|\theta_i,X)
    \end{aligned}
\end{equation}
where $p(\theta_i|X)$ indicates the probability of selecting expert $f_i$ given input $X$ and $p(Y|\theta_i,X)$ indicates the probability of predicting $Y$ given expert $f_i$ and input $X$.

\subsection{Basic Model: BOND}\label{sec:bond}
\begin{figure*}[t]
    \centering
    \includegraphics[width = \textwidth]{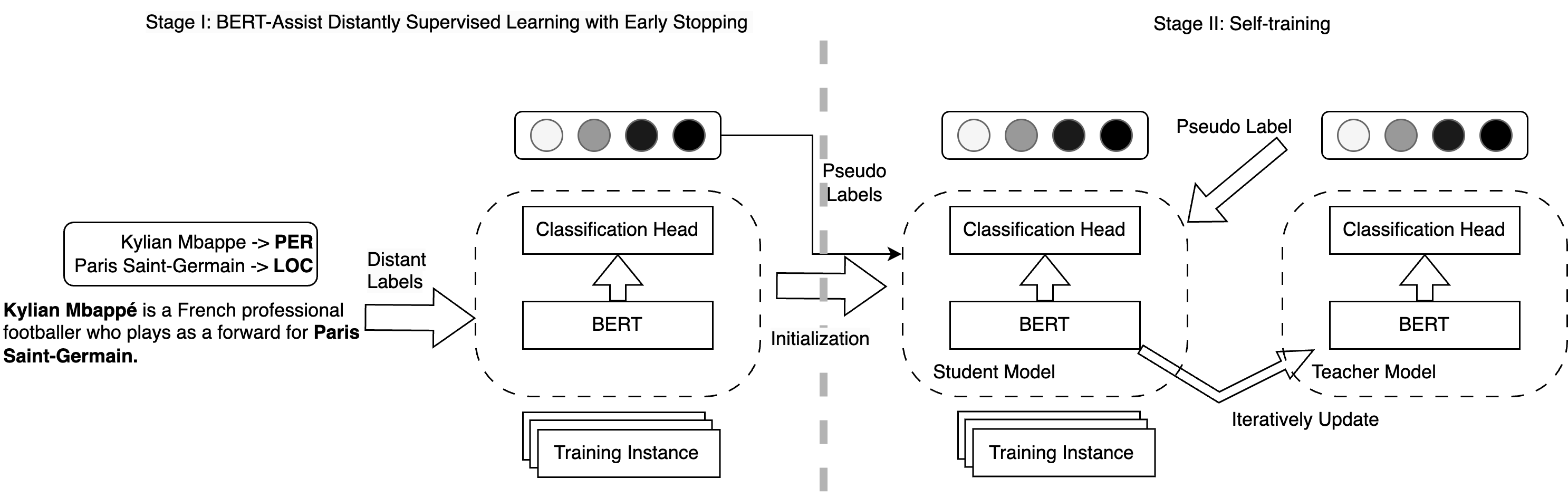}
    \caption{Overall framework of BOND~\cite{liang2020bond} including two stages: (1) distantly supervised NER based on BERT with early-stop, (2) self-training with pseudo labels via the student-teacher training paradigm.}
    \label{fig:bond}
\end{figure*}

BOND~\cite{liang2020bond} is a distantly supervised NER model based on BERT. As shown in Fig.~\ref{fig:bond}, the overall framework of BOND mainly includes two stages: (1) distantly supervised NER with BERT and (2) self-training with pseudo labels via the student-teacher training paradigm.

BOND is a distantly supervised NER model based on BERT, which includes two key stages. (1) distantly supervised NER with BERT (2) self-training with pseudo labels via the student-teacher training paradigm. In the first stage, we fine-tune a pretrained BERT model alongside an NER layer. This layer uses BERT’s token embeddings to predict entity labels for each token. Our approach operates on unlabeled documents and external knowledge bases. BOND identifies potential entities using POS tagging and hand-crafted rules, and then, the model querying the knowledge base for distant labels during training. To mitigate overfitting to noisy distant labels, BOND utilizes an early stopping strategy. In the second stage, we refine the model from step 1 by using a teacher-student training paradigm. The teacher generates pseudo labels for unlabeled documents, while the student optimizes its parameters using these labels. Through the process the model is fine-tuned the, resulting in improving distant supervision performance in NER.	

%% file: 4_Model.tex
\section{Model}

Our proposed model contains two components: (1) a Mixture of Experts (MoE) model trained with distantly supervised labels; (2) a self-training process with generated pseudo labels (similar to BOND). In this section, we first introduce the basic implementation of MoE model for NER task. Next, we propose a fair assignment algorithm to enhance the MoE's capability of training different experts. Last, we combine the MoE module into the BOND framework for the distantly supervised NER task.


\subsection{Mixture of Expert}

Mixture models~\cite{Shen2019MixtureMF} are latent variable models trained via EM algorithm. Given a token $x$ with its label $y$, a $K$-experts latent variable $z \in \{ 1, \dots, K\}$ stands for each expert, the decomposition of marginal likelihood is:

\begin{equation}
    \begin{aligned}
    p(z | x, y; \theta) = \frac{p(z|x; \theta) p(y|z, x; \theta)}{\sum_{z'} p(z' | x; \theta) p(y| z', x; \theta)} \\
    p(y | x; \theta) = \sum_{z=1}^K p(y, z| x; \theta) p(y | z, x; \theta)
    \end{aligned}
\end{equation}

By calculating the posterior probability $p(z | x, y; \theta)$ as the responsibilities of each expert with Bayes rule. The model could be trained by the hard-EM algorithms (a hard form of EM algorithm is applied because we want to maximaze the margin distribution between different experts):

\begin{equation}
    \begin{aligned}
    p(z | x, y; \theta) = \frac{p(z|x; \theta) p(y|z, x; \theta)}{\sum_{z'} p(z' | x; \theta) p(y| z', x; \theta)}
    \end{aligned}
\end{equation}

\begin{itemize}
    \item \textbf{E-step:} estimate the responsibilities of each expert $r_{z} \leftarrow [z = \arg \max_{z'} p(y, z'|x; \theta_{t})]$ using the current parameters $\theta_{t}$.
    \item \textbf{M-step:} update $\theta_{(t+1)}$ through each expert with gradients $\nabla_\theta \log p(y,z | x; \theta_{t})$ weighted by their responsibilities $r_z$.
\end{itemize}

To encourage all experts to generate good hypotheses for any source sentence, we set the prior $p(z|x; \theta)$ to be uniform. Which will turn the loss function into a simple form:

\begin{equation}
    \begin{gathered}
    \mathcal{L} (\theta) = \mathbb{E} [\min_z -\log p(y|z, x; \theta)]
    \end{gathered}
\end{equation}

\begin{figure}
    \centering
    \includegraphics[width = 0.48\textwidth]{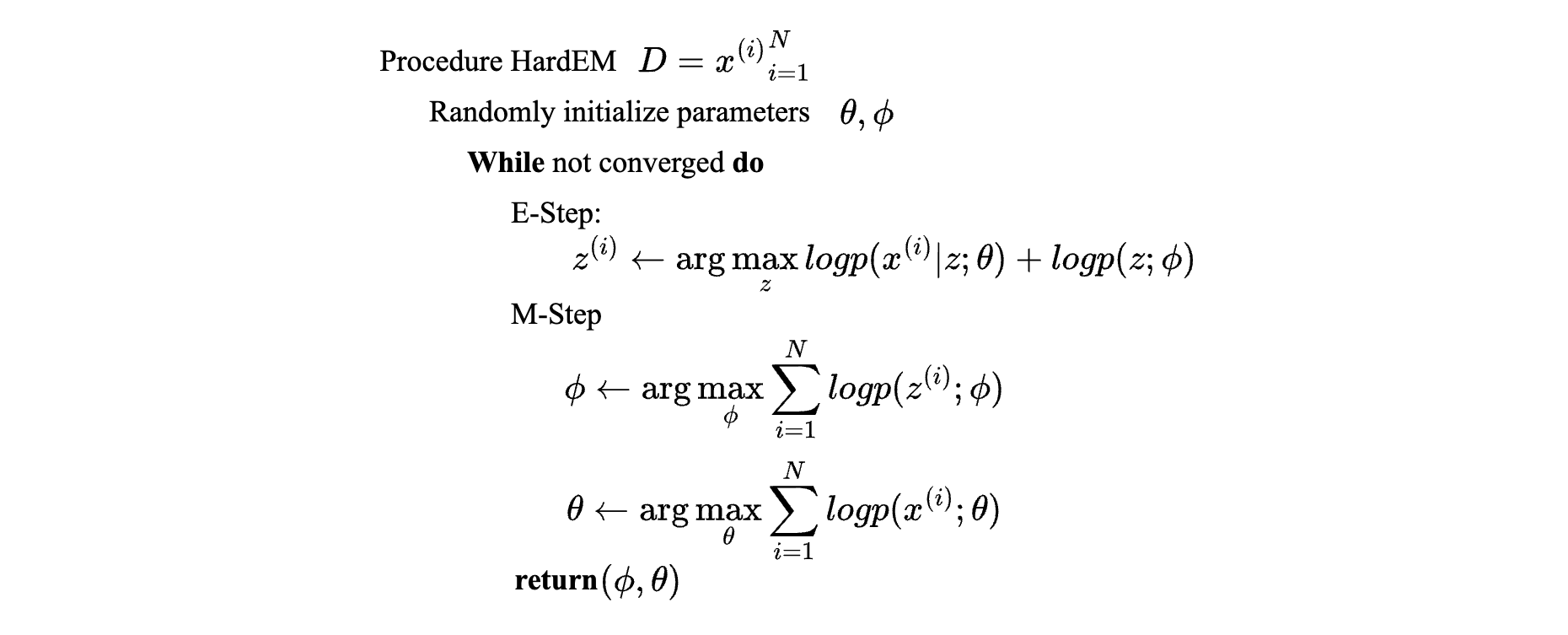}
    \caption{Mixture of Expert}
    \label{fig:moe}
\end{figure}

The pseduo code for the MoE module is shown in Fig.~\ref{fig:moe}. However, simply applying MoE into token-level NER may cause several problems, e.g., an entity can be split into several tokens and fed into different expert models, which will damage the information coherence. To solve this problem, we modified the naive implementation into document-level expert selection. That is, at E-step, we choose each expert with highest posterior probability on average scores among all tokens in each input document. During M-step, we update that expert by all the training examples in the assigned document.

\subsection{Fair Assignment}
The existing MoE framework, though promising, may faces an unfair assignment problem. Biased assignments may lead to extreme scenarios where all documents are assigned to the same model, while other models have no training data, which will dramatically degrade the MoE performance. To alleviate that issue, we introduce a fair assignment module atop MoE. In this model, we use a document-level fairness, treating documents as users and base models as utilities. The assignment process two-sided\cite{wu2021tfrom} fair recommendation, providing equitable item exposure and user utility. We address two-sided fairness by imposing marginal distribution constraints on the score matrix.

Specifically, the model set $M$ is represented as a discrete uniform distribution over each model, that is ${\mu}=\frac{\mathbf{1}_{|M|}}{|M|}$, and the document set $D$ is represented as a discrete uniform distribution over each document, that is ${\nu}=\frac{\mathbf{1}_{|D|}}{|D|}$. The marginal distribution of $\mathbf{S}$ along the document dimension $\sum_{d\in D}\mathbf{S}(d,m)$ corresponds to the exposure of model $m\in M$, and the marginal distribution of $\mathbf{S}$ along the model dimension $\sum_{m\in M}\mathbf{S}(d,m)$ corresponds to the document utility. To guarantee the two-sided assignment fairness, we pose two marginal distribution constraints as follows:

\begin{equation}
    \begin{aligned}
        \sum_{d\in D}\mathbf{S}(d,m) = \mu\\
        \sum_{m\in M}\mathbf{S}(d,m) = \nu\\
    \end{aligned}
\end{equation}

To satisfy the marginal constraints shown in Eq.~\eqref{eq:const}, we utilize the Sinkhorn matrix scaling technique~\cite{sinkhorn1967diagonal}. The idea is to compute a row scaling vector $\mathbf{a}$ and a column scaling vector $\mathbf{b}$ to scale the original $\mathbf{S}$. With randomly initialized vectors $\mathbf{a}^{(0)}$ and $\mathbf{b}^{(0)}$, the Sinkhorn algorithm iteratively updates $\mathbf{a}^{(t)},\mathbf{b}^{(t)}$ as follows.
\begin{equation}
    \begin{aligned}
        &\mathbf{a}^{(t)} = \frac{\mathbf{1}_{|D|}}{|D|\mathbf{S}\mathbf{b}^{(t-1)}}\\
        &\mathbf{b}^{(t)} = \frac{\mathbf{1}_{|M|}}{|M|\mathbf{S}^T\mathbf{a}^{(t)}}
    \end{aligned}
\end{equation}
And the two-sided fair score matrix can be computed as follows.
\begin{equation}
    \mathbf{S} = \text{diag}(\mathbf{a})\mathbf{S}\text{diag}(\mathbf{b})
\end{equation}

\subsection{BOND-MoE}
Combing the basic BOND framework with the MoE module and the fair assignment component, we have our BOND-MoE module. During training phase, we firstly provide a set of distantly labeled documents to a set of experts. The MoE module computes the posterior probability, assigning each document deterministically to the expert with the highest likelihood of accurate classification. Then each expert trains independently on its assigned dataset, ensuring uniqueness due to the hard EM process. Afterwards, all experts collaborate to re-label unlabeled documents. Their individual NER predictions are ensembled to mitigate noise. Our model then employs self-training to further boost the model performance. Compared with using documents labeled by a single model, combining the MoE expert reduce the noise in self-labeled data, especially during early training stages, where noisy labels may abound.

%% file: 5_Experiment.tex
\section{Experiments}\
\subsection{Experiment Setup}
\paragraph{Dataset Description.} We use 5 dataset for evaluation in this paper, including: (1) CoNLL03~\cite{sang2003introduction} which consists of 1393 English news articles with 4 entity types, (2) Twitter~\cite{godin2015multimedia} which consists of 2400 tweets with 10 entity types, (3) OntoNote5.0~\cite{AB2/MKJJ2R_2013} which consists of multi-domain articles with 18 entity types, (4) Wikigold~\cite{balasuriya2009named} which consists of Wikipedia articles with 4 entity types, and (5) Webpage~\cite{ratinov2009design} which consists of conference webpages with 4 entity types.

\paragraph{Baseline Methods} Five state-of-the-art distantly supervised NER methods are used as baseline methods in the experiments. BiLSTM-CRF~\cite{ma2016end} adopts a Bi-directional LSTM-based for token embedding and the CRF layer for token label prediction. AutoNER~\cite{shang2018learning} is built upon a fuzzy LSTM-CRF model, which is further trained via maximizing the overall likelihood. LRNT~\cite{cao2019low} applies partial-CRFs on high-quality data with non-entity sampling. ConNet~\cite{lan2019learning} utilizes multi-crowd annotation with dynamic aggregation based on attention mechanism. All these baseline methods utilize the same distant supervision from external knowledge base for training. We also compare with the baseline BOND~\cite{liang2020bond} method without MoE to show the improvement carried by our proposed fair-MoE module. 

\paragraph{Metrics.} In our experiments, we adopt precision, recall and $F_1$ score for evaluation.  F1 score is the harmonic mean of precision and recall, that is 

\begin{align}
    F_1=\frac{2\text{precision}*\text{recall}}{\text{precision}+\text{recall}}
\end{align}

\subsection{Result}

Our experiment results are shown in Table~\ref{tab:exp}. For all five benchmark datasets, our proposed model achieves best average results on $F_1$ score. It is shown that by incorporating our proposed MoE module, we can increase the performance from the baseline BOND framework. We summarize our findings as follows:

 Comparing with traditional methods (e.g., AutoNER~\cite{shang2018learning}),  pretrained language models-based motheds (e.g., BOND~\cite{liang2020bond}) achieve a much higher performance for distant NER task because of the implicit knowledge stored in PLM. 
    
Specifically, our proposed model outperforms the best tradition baseline method by up to 81.72\% in $F_1$ score, 16.16\% in precision and 142.00\% in recall. Compared with the basic BOND model, our proposed model further achieves an up to 1.58\% improvement in terms of the $F_1$ score, 5.40\% improvement in terms of precision and 1.40\% improvement in terms of recall.

In addition, as shown in Table~\ref{tab:exp}, our proposed BOND-MoE model outperforms the basic BOND model in most cases, which supports our claim that leverageing PLM for distant supervision may introduce noisy distant labels and the MoE module can help reduce such noisy labels, hence benefits the NER process. More detailed ablation study on the MoE module and the fair assignment module is carried out in the following section.

\begin{table}[t]
    \centering
    \small
    \scalebox{0.8}{
    \begin{tabular}{|c|c|c|c|c|c|}
    \hline
        Method &CoNLL03 &Tweet &OntoNote &WebPage &Wikigold \\\hline \hline
        BiLSTM-CRF &59.36 &21.69 &66.31 &43.45 &42.83 \\
        AutoNER &66.90 &25.90 &66.58&51.40 &47.46 \\
        LRNT &67.94 &23.64 &67.49 &47.83&46.12 \\
        ConNET &74.23 &- &- &- &- \\
        BOND &77.12 &\textbf{48.07} &71.09 &63.69 &55.76 \\
        Ours &\textbf{79.31} &47.76&\textbf{71.00} &\textbf{64.11} &\textbf{56.59} \\ \hline
    \end{tabular}}
    \caption{Results on Real-world Datasets(F1 score)}
    \label{tab:exp}
\end{table}


To better understand each component of our proposed framework, we also conduct ablation study by taking off each module to measure the contribution of each module to the final result respectively. We note that by dropping the MoE module, the model degenerates to the original BOND framework. The experiment results are shown in Tab.~\ref{tab:abla}. First, the whole model with all three stages yields the best performance which proves the success of our proposal, achieving up to 6.64\% improvement in terms of $F_1$ score, 10.05\% improvement in terms of precision and 10.90\% improvement in terms of recall. Next, using the self-training framework shows a large increase in the $F_1$ score than simply fine tuning the PLM, indicating that the self-supervised training process can improve the model fitting and reduce the noise of the pseudo-labels~\cite{liang2020bond}. Besides, it is shown that the MoE module can consistently improve the model precision, this is consistent with our assumption that distant labels are noisy and incorporating MoE can reduce such noise, hence help train a more robust and precise NER model. In addition, it is shown that without the fair assignment algorithm, the model's performance will be damaged due to the baised training problem.

\begin{table}[t]
    \centering
    \small
    \scalebox{0.8}{
    \begin{tabular}{|l|c|c|c|c|c|}
    \hline
        Method &CoNLL03 &Tweet &OntoNote &WebPage &Wikigold \\\hline \hline
        All Stages &\textbf{79.43} &47.54 &\textbf{71.12} &\textbf{65.02} &\textbf{56.27} \\ \hline
        w/o MoE &78.45 &\textbf{47.87} &70.23&63.23 &54.90 \\ \hline
        w/o Fair-Assign &78.67 &47.36 &69.09&62.89 &55.30 \\ \hline
        w/o Self-train &74.21 &46.25 &67.46 &63.64 &57.24 \\ \hline
    \end{tabular}}
    \caption{Results on taking off different components in our model (F1 score)}
    \label{tab:abla}
\end{table}

%% file: 6_Conclusion.tex
\section{Conclusion}
In this paper, we proposed BOND-MoE, a novel framework that combines pretrained language models with a MoE structure for distantly supervised NER. Our method capitalizes on the strengths of both pretrained language models and expert-based architectures. we train individual experts on distinct input documents using a hard-EM algorithm. Each expert focuses on diverse categories within the same named entity, resulting in a more robust model that is less affected by noisy annotations and label matching ambiguities. We also propose two marginal distribution constraints based on matrix scaling techniques, ensuring fair allocation of documents to each expert. Additionally, we employ a self-training process, updating model parameters and generating pseudo labels for subsequent training steps. Experimental results on five real-world datasets demonstrate that BOND-MoE outperforms baseline models in terms of F1 score, highlighting the effectiveness of incorporating pretrained language models within the MoE structure for distant supervision in NER.

Our proposed model also has some limitations, it sometimes classifies the phrases entities as separate words. In the future, we plan to explore more distantly supervised models and extend our MoE approach to tackle related tasks such as relation extraction and event discovery